\documentclass[runningheads]{llncs}
\usepackage{graphicx}
\usepackage{comment}
\usepackage{amsmath,amssymb} 
\usepackage{color}

\usepackage{cuted}
\usepackage{capt-of,etoolbox}

\usepackage{epsfig}
\usepackage{graphicx}
\usepackage{bm}
\usepackage{multirow}
\usepackage{hyperref}

\newcommand{\etal}{{\em et al.}}

\newcommand{\tref}[1]{Tab.~\ref{#1}}

\newcommand{\fref}[1]{Fig.~\ref{#1}}
\newcommand{\sref}[1]{Sec.~\ref{#1}}

\newcommand{\tabincell}[2]{\begin{tabular}{@{}#1@{}}#2\end{tabular}}

\begin{document}
\pagestyle{headings}
\mainmatter
\def\ECCVSubNumber{961}  

\title{Pyramid Multi-view Stereo Net with Self-adaptive View Aggregation} 

\titlerunning{PVA-MVSNet}
%

\makeatletter
\newcommand\footnoteref[1]{\protected@xdef\@thefnmark{\ref{#1}}\@footnotemark}
\makeatother

\def\thefootnote{*}\footnotetext{Equal Contribution}
\def\thefootnote{$\dagger$}\footnotetext{Corresponding Author}

\author{Hongwei Yi\inst{1}$^*$ \and
Zizhuang Wei\inst{1}$^*$ \and
Mingyu Ding\inst{2} \and Runze Zhang\inst{3}$^\dagger$ \and 
Yisong Chen\inst{1} \and Guoping Wang\inst{1} \and
Yu-Wing Tai\inst{4}
}
\authorrunning{H. Yi, Z. Wei et al.}
%
\institute{
PKU \hspace{2mm}
\email{\{hongweiyi, weizizhuang, chenyisong, wgp\}@pku.edu.cn} \and
HKU \hspace{2mm} \email{myding@cs.hku.hk} \and
Tencent \hspace{2mm}
\email{ryanrzzhang@tencent.com} \and
Kwai Inc. \hspace{2mm}
\email{yuwing@gmail.com}
}
\maketitle

\vspace{-3mm}
\begin{abstract}
In this paper, we propose an effective and efficient pyramid multi-view stereo (MVS) net with self-adaptive view aggregation for accurate and complete dense point cloud reconstruction.
Different from using mean square variance to generate cost volume in previous deep-learning based MVS methods, 
our \textbf{VA-MVSNet} incorporates the cost variances in different views with small extra memory consumption by introducing two novel self-adaptive view aggregations: pixel-wise view aggregation and voxel-wise view aggregation.
To further boost the robustness and completeness of 3D point cloud reconstruction, we extend VA-MVSNet with pyramid multi-scale images input as \textbf{PVA-MVSNet}, where multi-metric constraints are leveraged to aggregate the reliable depth estimation at the coarser scale to fill in the mismatched regions at the finer scale. 
Experimental results show that our approach establishes a new state-of-the-art on the \textsl{\textbf{DTU}} dataset with significant improvements in the completeness and overall quality, and has strong generalization by achieving a comparable performance as the state-of-the-art methods on the \textsl{\textbf{Tanks and Temples}} benchmark. 
Our codebase is at \hyperlink{https://github.com/yhw-yhw/PVAMVSNet}{https://github.com/yhw-yhw/PVAMVSNet}
\end{abstract}

\section{Introduction}
Multi-view Stereo (MVS) aims to recover dense 3D representation of scenes using stereo correspondences as the main cue given multiple calibrated images~\cite{schonberger2016structure,lhuillier2005quasi,seitz2006comparison,strecha2008benchmarking}.
Although they have achieved great success on MVS benchmarks~\cite{aanaes2016DTU,knapitsch2017tanks,schops2017ETH3D}, many of them still have limitations in handling matching ambiguity and usually have a low completeness of 3D reconstruction. 
Recently, the deep neural network has made tremendous progress in multi-view stereo~\cite{huang2018deepmvs,yao2018mvsnet,im2019dpsnet}. These methods learn and infer the information hardly obtained by stereo correspondences in order to handle matching ambiguity. However, they do not learn and utilize the following important information.

\begin{figure}[t]
\centering
\includegraphics[width=0.85\columnwidth]{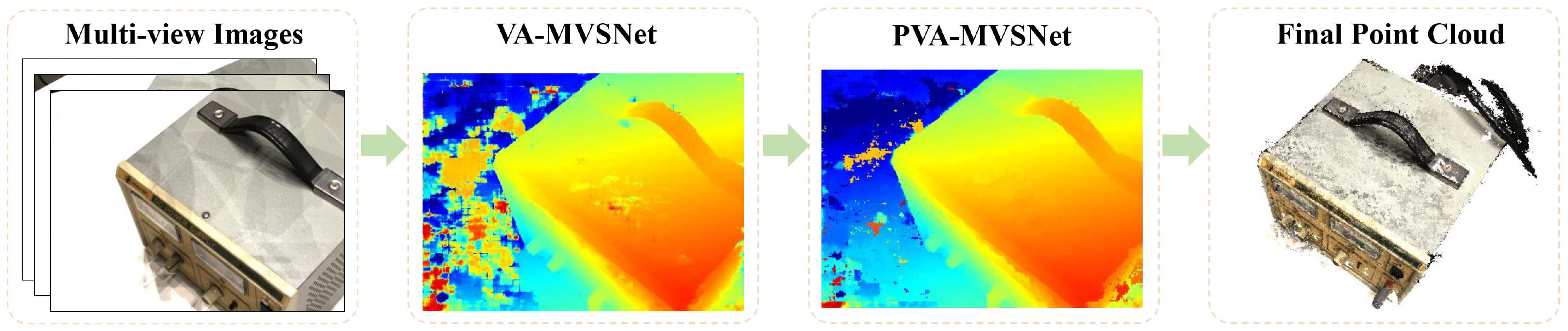}
\vspace{-3mm}
\caption{VA-MVSNet performs an efficient and effective multi-view stereo with self-adaptive view aggregation to generate an accurate depth map. Cast in pyramid images additionally, PVA-MVSNet aggregates multi-scale depth maps with multi-metric constraints to boost the point cloud reconstruction with high accuracy and completeness.}
\label{fig_teaser}
\vspace{-5mm}
\end{figure}

First, the one-stage end-to-end deep MVS architectures~\cite{yao2018mvsnet,yao2019recurrent,im2019dpsnet} that directly learn from images all follow the philosophy that all view images contribute equally to the matching cost volume~\cite{hartmann2017learned}. For instance, MVSNet~\cite{yao2018mvsnet} and R-MVSNet~\cite{yao2019recurrent} both apply the mean square variance operation on multiple cost volumes, and DPSNet~\cite{im2019dpsnet} selects the mean average operation. However, images from different views lead to heterogeneous image capture characteristics due to different illumination, camera geometric parameters, scene content variability, etc. Based on this observation, we propose a self-adaptive view aggregation module to learn the different significance in multiple matching volumes among images from different views. Our module benefits from the aggregated features by a self-adaptive fusion, where better element-wise matched regions are enhanced while the mismatched ones suppressed.

Second, the multi-scale information is not leveraged well to improve the robustness and completeness of 3D reconstruction. Unlike ACMM~\cite{Xu_2019_CVPR} where pyramid images are processed progressively to regress the depth map in a coarse-to-fine manner, we propose a novel way to aggregate multi-scale pyramid depth maps which are generated in parallel by multi-metric constraints to a refine depth map.
In particular, to correct the mismatched regions at the finer depth map, we progressively aggregate the reliable depth at the coarser level to refine the finer depth map but do not introduce quantization errors benefiting from our multi-metric constraints.

To this end, we propose a novel efficient and effective pyramid multi-view stereo network with self-adaptive view aggregation, denoted as \textbf{PVA-MVSNet}.
Our method constructs multi-scale pyramid images and processes them in parallel by \textbf{VA-MVSNet} to produce pyramid depth maps. 
To regularize 3D warping feature volumes from different views, we propose two self-adaptive element-wise view aggregation modules to learn different variance of different views in an order-independent manner. Through a depth map estimator, 3D cost volume is utilized to estimate the corresponding depth map.
To further improve the robustness and completeness of 3D reconstruction generated by VA-MVSNet,
our proposed multi-metric pyramid depth aggregation corrects the mismatched regions at finer depth maps using the reliable depths at coarser depth maps by checking photometric and geometric consistency. 

Our main contributions are listed below:
\begin{itemize}
\item We propose self-adaptive view aggregation to incorporate the element-wise variances among images from different views, guiding the multiple cost volumes to aggregate a normalized one.
\item We investigate to incorporate multi-scale information by our multi-metric pyramid depth maps aggregation in PVA-MVSNet, to further improve the robustness and completeness of 3D reconstruction.
\item Our method establishes a new state-of-the-art on the \textsl{DTU} and a comparable performance as the state-of-the-art methods on the \textsl{Tanks and Temples}.
\end{itemize}


\section{Related Work}

\paragraph{Traditional MVS Reconstruction:} 
Traditional MVS reconstruction algorithms can be divided into four types: voxel based~\cite{sinha2007multi,vogiatzis2007multiview},
surface based~\cite{hiep2009towards,cremers2010multiview},
patch based~\cite{goesele2007multi,furukawa2009accurate} and depth map based methods~\cite{furukawa2009accurate,campbell2008using,tola2012efficient,galliani2015massively,Zheng_2014_CVPR,schonberger2016pixelwise}.
Among those methods, the depth map based approaches are more concise and flexible. 
Recently, many advanced MVS algorithms estimate high quality depth maps by view selection, local propagation and multi-scale aggregation strategies.
Zheng~\etal~\cite{Zheng_2014_CVPR} propose a depth map estimation method by solving a probabilistic graphical model. Schönberger~\etal~\cite{schonberger2016pixelwise} present a new MVS system named COLMAP 
where geometric priors are used to better depict the probability of their graphical model. Xu~\etal~\cite{xu2019multi} propose a multi-scale MVS framework with adaptive checkerboard propagation and multi-hypothesis joint view selection to improve the performance.
These works utilize predefined criteria for pixel-wise view selection, which cannot be adaptive for different scenes.

\paragraph{Learning Based Stereo Matching:}
Recently, the convolutional neural network (CNN) has made tremendous progress in many vision tasks~\cite{kendall2017end,dosovitskiy2015flownet,yang2018segstereo,ren2015faster,song2018edgestereo}, including several attempts on multi-view stereo.
Early learning-based methods~\cite{huang2018deepmvs,flynn2016deepstereo,ji2017surfacenet} pre-warp the images to generate plane-sweep volumes as the input.
Two promising approaches~\cite{yao2018mvsnet,im2019dpsnet} both propose the differential homography warping, which implicitly encodes multi-view camera geometries into the network and enables an end-to-end training fashion. 
Furthermore, R-MVSNet~\cite{yao2019recurrent} replaces 3D-CNN in MVSNet~\cite{yao2019recurrent} by the gated recurrent unit (GRU) to reduce memory consumption during the inference phase.
Gu et al.~\cite{Gu_2020_CVPR} and Cheng et al.~\cite{Cheng_2020_CVPR} both propose a cascaded MVS network through constructing coarse-to-fine cost volume which eases the memory limitation of the volume resolution in comparison with uniformly sampled cost volume~\cite{yao2018mvsnet,yao2019recurrent,ChenPMVSNet2019ICCV}.
P-MVSNet~\cite{luo2019p} proposes a patch-wise matching module to learn the isotropic matching confidence inside the cost volume.
Particularly, those methods follow the philosophy that the feature volumes from different view images contribute equally, neglecting heterogeneous image capturing characteristics due to different illumination, camera geometric parameters and scene content variability.
PointMVSNet~\cite{chen2019point} is a two-stage coarse-to-fine method which needs a coarse depth map by a lower-resolution version MVSNet~\cite{yao2018mvsnet}.

Based on the above analysis, we propose a self-adaptive view aggregation module to incorporate the different significance in multiple feature volumes from different views, where better element-wise matched features can be enhanced while the mismatched errors can be suppressed. 
To further improve the robustness and completeness of 3D reconstruction point cloud, we propose a multi-metric pyramid depth aggregation to aggregate multi-scale information in pyramid images. The mismatched depth value generated by the original image can be filled-in by the reliable depth value from the downsized image under photometric and geometric consistency.

\begin{figure}[t]
  \centering
  \includegraphics[width=0.95\columnwidth]{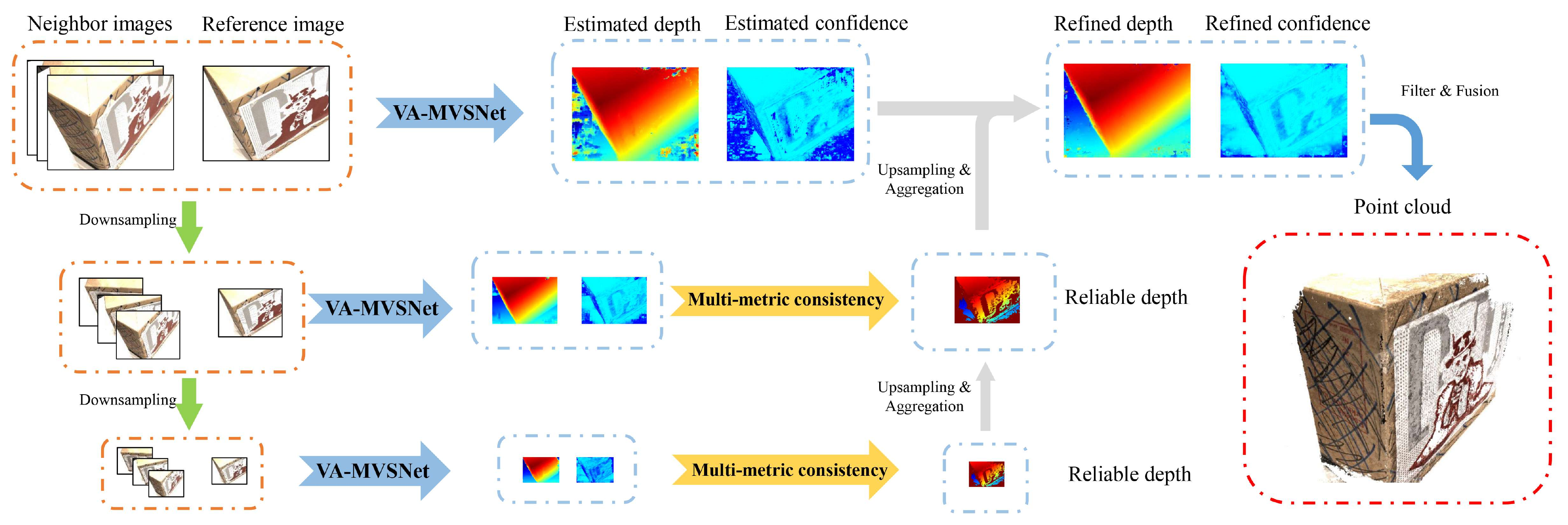}
  \caption{Overview of PVA-MVSNet. We firstly input multi-scale pyramid images to VA-MVSNet to generate corresponding pyramid depth maps in parallel. Then we progressively replace the mismatched depths in the finer depth map with more reliable depths from a coarser level to achieve a refined depth map. Finally, we reconstruct the point cloud by filter and fusion through all estimated depth maps of the image set.}
  \label{overall}
\end{figure}

\section{Method}
We first describe the overall architecture of PVA-MVSNet in \sref{soverall}. Then, we introduce the details of VA-MVSNet in \sref{sva}. Finally, we present the multi-metric pyramid depth aggregation in \sref{mmp}.

\subsection{Overall}\label{soverall}
Given a reference image $\boldsymbol{I}_{i=0}$ and $\boldsymbol{I}_{i=1,\cdots, N-1}$ neighboring images and corresponding calibrated camera parameters $\boldsymbol{Q}_{i=0}$ and $\boldsymbol{Q}_{i=1,\cdots, N-1}$, where $N$ represents the number of multi-view images, our goal is to estimate the depth map for each reference image. Afterwards, we filter and fuse all estimated depth maps to reconstruct 3D point clouds. 

For the depth estimation of a reference image, our main architecture is illustrated in~\fref{overall}. We construct an image pyramid with $K$ multiple scales for all images with a downsampling scale factor $\eta$. We denote $k$-level pyramid images and corresponding camera parameters as $I_{i=0,\cdots,N-1}^{k}$ and $Q_{i=0,\cdots,N-1}^{k}$ respectively, where $k=0,\cdots,K-1$. The scale $k=0$ of the pyramid represents the original image. 
We process each level images in the pyramid by VA-MVSNet to obtain depth maps of different scales in parallel. 
Then we progressively propagate the reliable depths from images with the lower resolution, which satisfy multi-metric constraints, to correct the mismatched errors of images with the higher resolution by replacements.
Finally, we obtain the refined depth map of the raw image. We term our whole method PVA-MVSNet.
\begin{figure}[t]
  \centering
  \includegraphics[width=0.95\columnwidth]{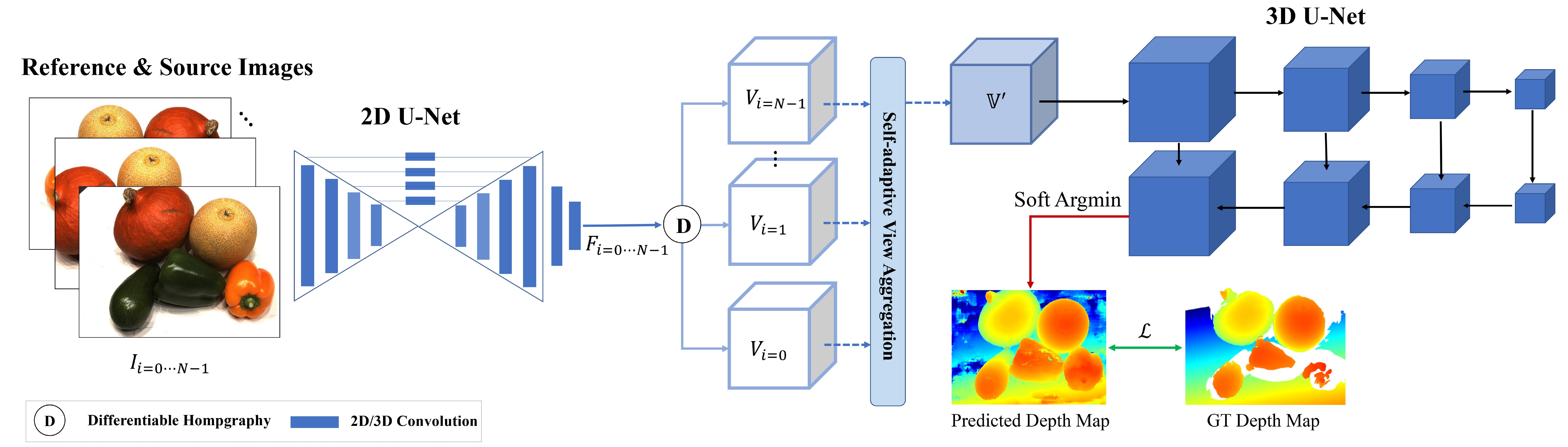}
  \caption{The network architecture of VA-MVSNet. Multi-view images go through 2D U-Net and differentiable homography warping to generate 3D feature volumes. Cost variances in different views are encoded in self-adaptive view aggregation to aggregate the 3D cost volumes which is regularized by 3D U-Net to regress the depth map.}
  \label{architecture}
\end{figure}

\subsection{Self-adaptive View Aggregation}\label{sva}
In VA-MVSNet in~\fref{architecture}, we first design a 2D U-Net to extract $\left\{\boldsymbol{F}_{i}\right\}_{i=0}^{N-1}$ feature maps with larger receptive fields from the $N$ input images.
For efficient computation, the output feature map is downsampled by four to the original image size with $32$ channels.

\begin{figure}[t]
    \centering
    \includegraphics[width=0.6\columnwidth]{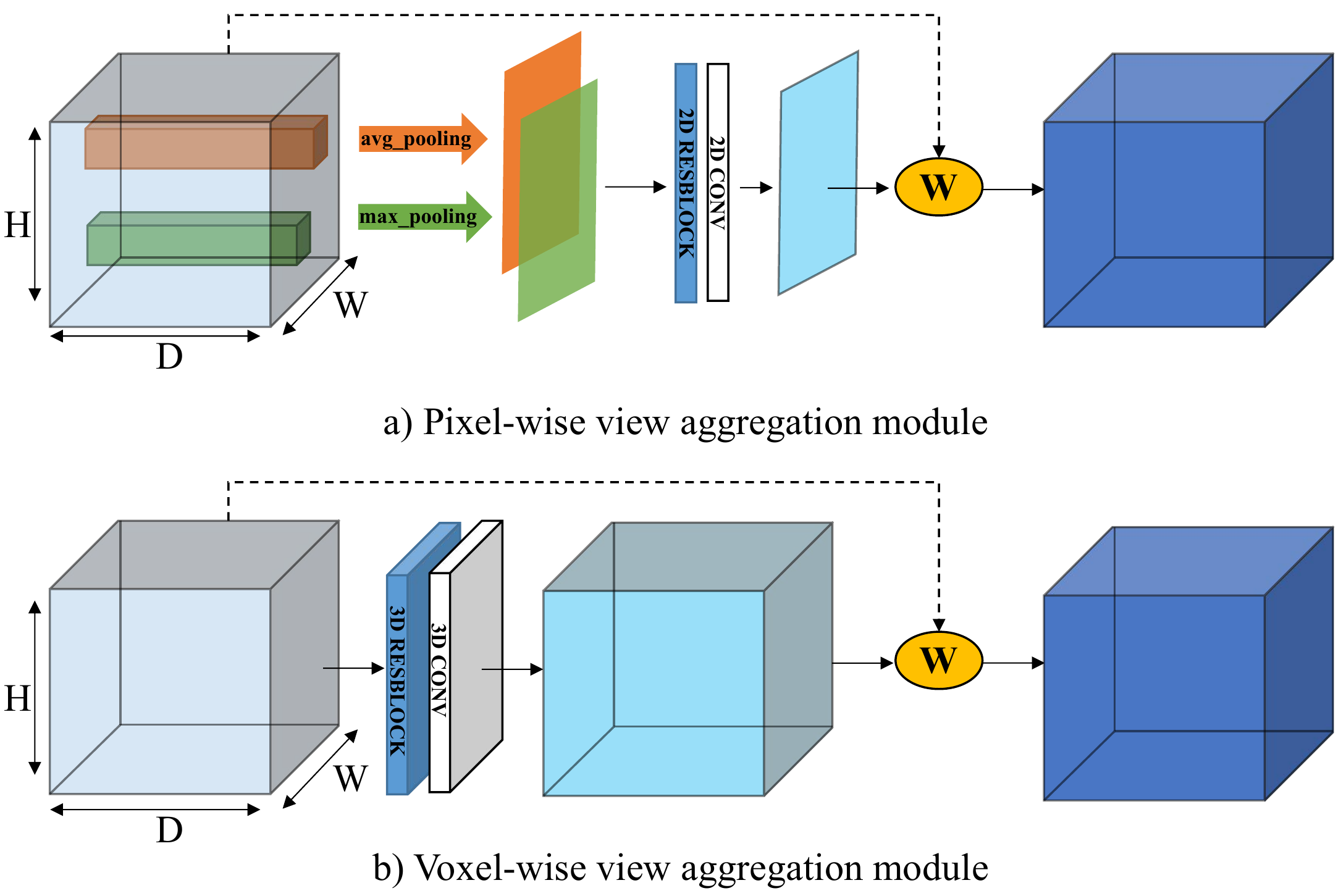}
    \caption{Illustration of two different self-adaptive element-wise view aggregation modules, a) pixel-wise view aggregation, b) voxel-wise view aggregation.}
    \label{pwva}
\end{figure}

Then each feature map from different views will be warped to the reference camera frustum by the differential homography~\cite{yao2018mvsnet,im2019dpsnet} with sampling $D_{i}$ layers to build 3D plane-sweep feature volumes $\boldsymbol{V}_{i}$.
To handle arbitrary $N$-view images input and the variances among images from different sources, we propose self-adaptive view aggregation to merge $\boldsymbol{V}_{i=0,\cdots,N-1}$ 3D feature volumes into one cost volume $\boldsymbol{C}$. 
Let $W_{i}, H_{i}, D_{i}, C_{i}$ denote the width, height, depth sample number and channel number of the input 3D warping feature volume from image $i$ respectively, the feature volume size can be represented as $S_{i}=W_{i}\cdot H_{i}\cdot D_{i}\cdot C_{i}$, the cost volume $C_{i}$ aggregation can be defined as a function:
$M:\underbrace{\mathbb{R}^{S_{i}} \times \cdots \times \mathbb{R}^{S_{N-1}}}_{N} \rightarrow \mathbb{R}^{S}$.
In previous work \cite{yao2018mvsnet,yao2019recurrent,im2019dpsnet}, 
this is a constant function where all views contribute equally, which is the mean square error of all input feature volumes. However, it is not reasonable due to different illumination, camera position, occlusion and image content etc, where a near reference image with no occlusion can provide more accurate geometric and photometric information than a far one with partial occlusion. 
Thus, we propose to employ self-adaptive view aggregation as this function to flexibly learn the potential different view variance from training data.
To achieve this goal, we develop and investigate two different self-adaptive view aggregation modules in \fref{pwva}, which shows how the self-adaptive view selection incorporates the variance between different views. 
We introduce the attention mechanism~\cite{vaswani2017attention,xu2015show} for guiding the network to select important matching information in different views.
In the point-wise view selection, similar as ACMM~\cite{Xu_2019_CVPR}, we consider that each pixel in the height and width dimension of 3D cost volume has different saliency but is consistent in the depth dimension. The voxel-wise view selection module is a 3D attention-guided mechanism to guide each voxel in 3D feature volumes to learn its own weight.

\paragraph{Pixel-wise View Aggregation.}
The pixel-wise view aggregation introduces a selective weighted attention map in the height and width dimension which considers the depth number hypothesis sharing common focusing weight. Given multi-view feature volumes $V_{i=0\cdots N-1}$, our regularized cost volumes are aggregated as $\boldsymbol{c}_{d,h,w}$:
\begin{equation}
     \boldsymbol{v'}_{i,d,h,w}=\boldsymbol{v}_{i,d,h,w}-\boldsymbol{v}_{0,d,h,w}, 
\end{equation}
\begin{equation}
     \boldsymbol{c}_{d,h,w}=\frac{\sum_{i=1}^{N-1}(1+\boldsymbol{w}_{h,w})\odot\boldsymbol{v'}_{i,d,h,w}}{N-1},
\end{equation}
where $\boldsymbol{w}_{h,w}$ represents a 2D weighted attention map to encode the various pixel-wise saliency among images from different sources and the reference view, and $\odot$ represents element-wise multiply operation. 

To generate a 2D weighted attention map, we design a \textsl{PA-Net} in \tref{weightnet} which consists of several 2D convolutional filters and a ResNet block~\cite{he2016deep} with the squeezing 2D features from $V'_{i}$ as input to learn the $\boldsymbol{w}_{h,w}$:
\begin{equation}
    \boldsymbol{w}_{h,w}=\textsl{PA-Net}(\boldsymbol{f}_{h,w}),
\end{equation}
\begin{equation}
    \boldsymbol{f}_{h,w} = \textsl{CONCAT}(\textsl{max\_pooling}(\left\|\boldsymbol{v'}_{d,h,w}\right\|_{1}), \textsl{avg\_pooling}(\left\|\boldsymbol{v'}_{d,h,w}\right\|_{1})), 
\end{equation}
where both $\textsl{max\_pooling}$ and $\textsl{avg\_pooling}$ are used to extract the highest and average cost matching information in the \textsl{depth} dimension, and $\textsl{CONCAT}(\cdot)$ denotes the concatenation operation.

\paragraph{Voxel-wise View Aggregation.}
The voxel-wise view aggregation module considers that each pixel with different depth layer hypothesis $d$ is treated differently, where each voxel in 3D feature volume learns its own importance. Based on this, we design a \textsl{VA-Net} as shown in \tref{weightnet} to directly learn the 3D weighted attention map with 3D convolutional filters for selecting useful cost information. The regularized 3D cost volumes $\boldsymbol{c}_{d,h,w}$ are aggregated by $\boldsymbol{v'}_{i,d,h,w}$:
\begin{equation}
     \boldsymbol{c}_{d,h,w}=\frac{\sum_{i=1}^{N-1}(1+\boldsymbol{w}_{d,h,w}) \odot \boldsymbol{v'}_{i,d,h,w}}{N-1}.
\end{equation}

\subsection{Depth Map Estimator}\label{ctf}
We design a 3D convolutional U-Net by leveraging different level information and expanding receptive fields to generate the probability volume $P$ with a \textsl{softmax} operation along the depth dimension. The details of 3D U-Net are in the supplementary material.

\begin{figure}[t]
    \centering
    \includegraphics[width=1.0\columnwidth]{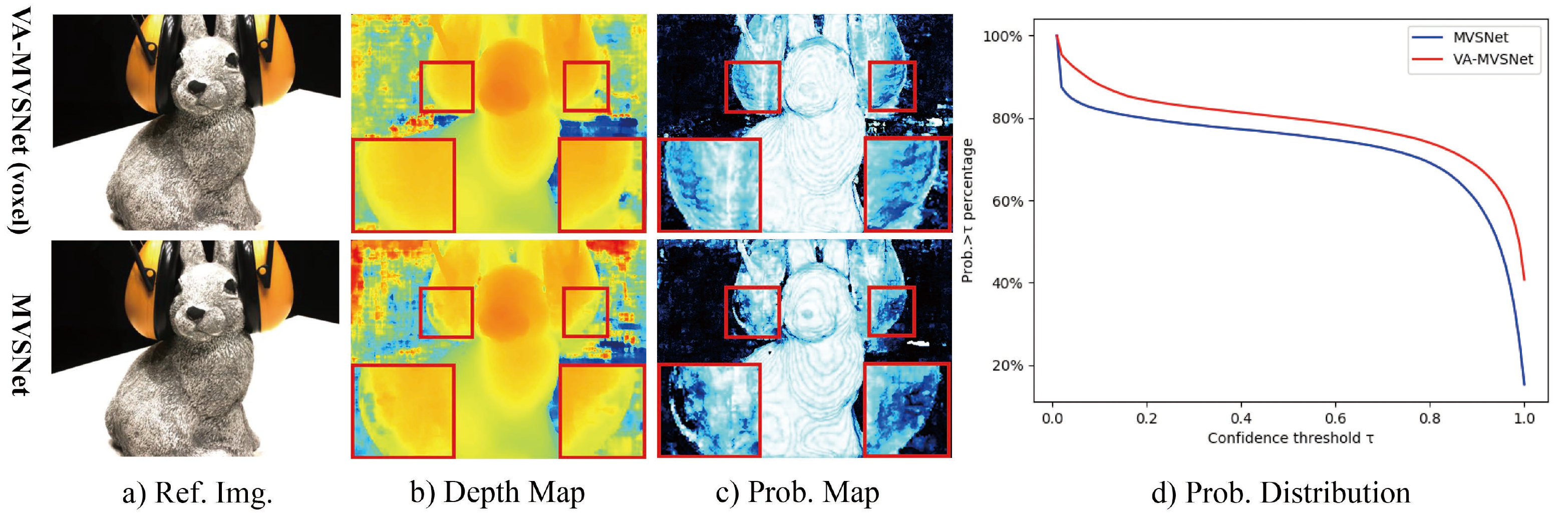}
    \caption{Comparison on the regressed depth map, probability map and probability distribution with MVSNet~\cite{yao2018mvsnet}. (a) One reference image of Scan $12$; (b) the inferred depth map; (c) the probability map; (d) the distribution of the probability map. Our self-adaptive view aggregation enhances the multi-view stereo network to generate more delicate and accurate depth estimations with higher confidence.}
    \label{conf_dist}
\end{figure}

To produce a continuous depth estimation, we use \textsl{soft argmin} operation~\cite{honari2018improving} on the output probability volume $P$ to estimate the depth $\boldsymbol{E}$:
\begin{equation}
\boldsymbol{E}=\sum_{d=d_{\min }}^{d_{\max }} d \times \boldsymbol{P}(d),
\end{equation}
where $\boldsymbol{P}(d)$ denotes the estimated probability of all pixels for the depth hypothesis $d$. 
Following MVSNet~\cite{yao2018mvsnet}, the probability map is calculated by the sum over the nearest four hypotheses in the 3D probability volume to measure the estimation quality.
Comparing the estimated depth map and confidence map in \fref{conf_dist} with \cite{yao2018mvsnet}, VA-MVSNet generates more reliable and accurate depth map with higher confidence benefiting from self-adaptive view aggregation. 

\begin{table}[htbp]
    \centering
    \scriptsize
	\begin{tabular}{c|c|c|c}
    \hline
    Input & Layer & Output & Output Size \\ \hline \hline
    \multicolumn{4}{c}{\textsl{PA-Net}} \\ \hline \hline
   $\boldsymbol{f}_{h,w}$ &  ConvGR,K=3,S=1,F=16 & wc\_0 & $W \times H \times 16$ \\ 
    wc\_0  &  ResBlockGR,K=3,S=1,F=16 & wres\_1 & $W \times H \times 16$ \\ 
    wres\_1  &  Conv,K=3,S=1,F=1 & wc\_2 & $W \times H \times 1$ \\
    wc\_2  &  Sigmoid & weight & $W \times H \times 1$ \\ \hline \hline
    \multicolumn{4}{c}{\textsl{VA-Net}} \\ \hline \hline
    $\boldsymbol{v^{'}}$ &  Conv3DGR,K=3,S=1,F=1 & wc3d\_0 & $D \times W \times H \times 1$ \\ 
    wc3d\_0 &  Conv3D,K=3,S=1,F=1 & wc3d\_1 & $D \times  W \times H \times 1$ \\ 
    wc3d\_1 &  Sigmoid & weight3d & $D \times  W \times H \times 1$ \\ \hline \hline
	\end{tabular}
	\caption{The details of \textsl{PA-Net} and \textsl{VA-Net}. We denote Conv, Conv3D as 2D and 3D convolution respectively, and use GR to represent the abbreviation of group normalization and the Relu. + and \& represent the element-wise addition and concatenation. K, S, F are the kernel size, stride and output channel number. N, H, W, D denote input view number, image height, image height and depth hypothesis number.}
	\label{weightnet}
\end{table}

\paragraph{Training Loss}
We use the same training losses in MVSNet~\cite{yao2018mvsnet}, which is the mean absolute error defined as $\mathcal{L}$:
\begin{equation}
\mathcal{L}=\sum_{\boldsymbol{x} \in \boldsymbol{x}_{valid}}\left\|\boldsymbol{d}(\boldsymbol{x})-\boldsymbol{\hat{d}}(\boldsymbol{x})\right\|_{1},
\end{equation}
where $\boldsymbol{x}_{valid}$ denotes the set of valid pixels in the ground truth, $\boldsymbol{d}(\boldsymbol{x})$ and $\boldsymbol{\hat{d}}(\boldsymbol{x})$ represent the estimated depth map and the ground truth respectively.

\subsection{Multi-metric Pyramid Depth Map Aggregation}\label{mmp}

So far, our proposed network VA-MVSNet generates good-enough depth maps for the point cloud reconstruction. To further improve the robustness and completeness of 3D reconstruction, we propose a novel multi-metric pyramid depth aggregation to aggregate reliable depth estimations in a lower-resolution depth map into a higher-resolution depth map, by replacing corresponding mismatched errors.

\begin{figure}[t]
    \centering
    \includegraphics[width=0.6\columnwidth]{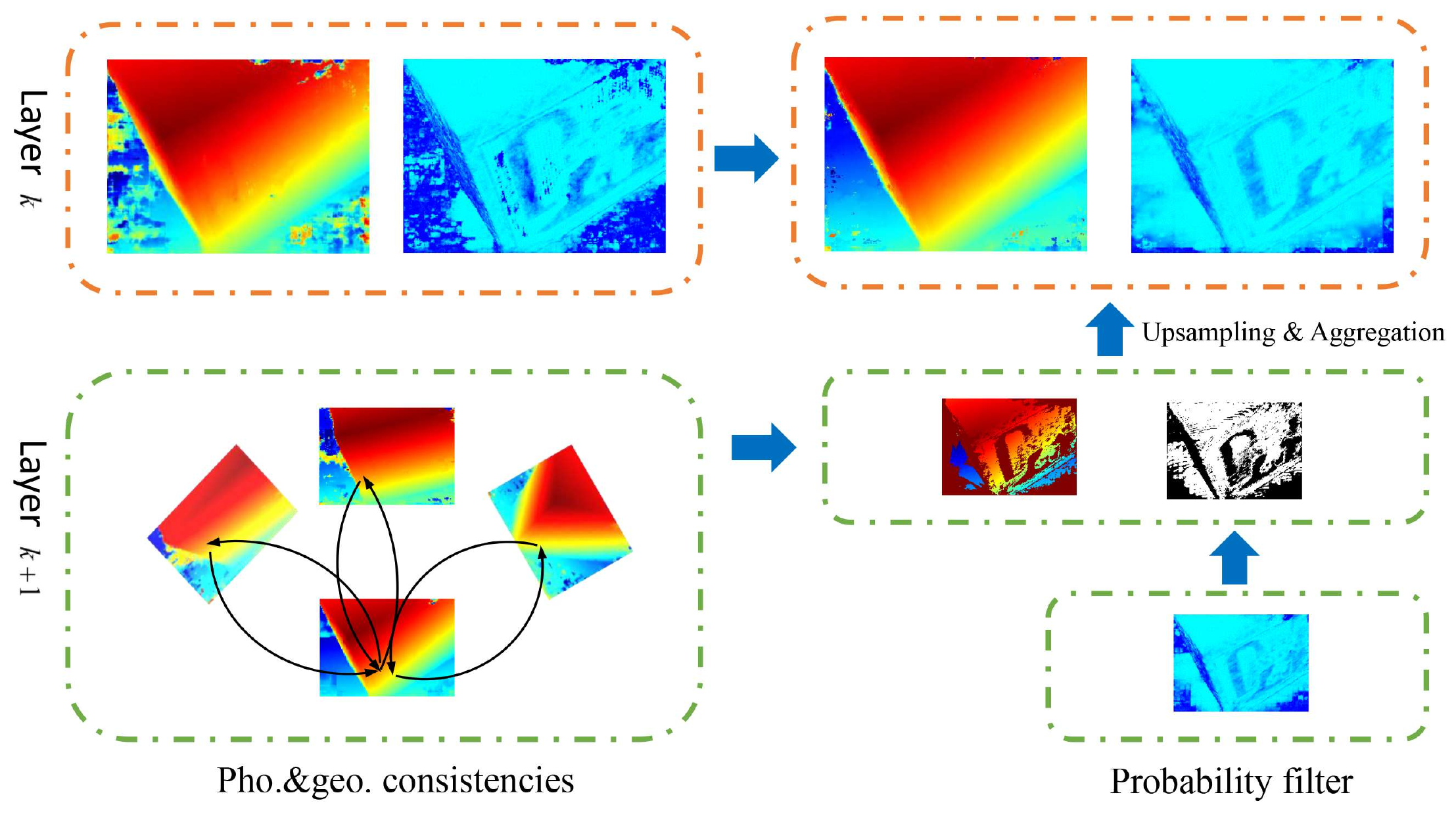}
    \caption{Illustration of multi-metric pyramid depth map aggregation, where the reliable depth at a lower scale level $k+1$ selected by multi-metric constraints, are used to fill-in the mismatched errors at a higher scale level $k$ by upsampling and aggregation.}
    \label{Multi-metric}
\end{figure}

In a higher-resolution fine estimated depth map, there are still some inaccurate depths with low confidences due to the matching ambiguity.
Note that the same convolutional filter generally extracts less local-wise, but more global information due to a larger receptive field from a downsampled image in comparison to the original image.
Quite different from ACMM~\cite{Xu_2019_CVPR}, which casts a image pyramid into VA-MVSNet to generate multi-scale depth maps in parallel, we propose to utilize multi-metric constraints, specifically, geometric and photometric consistency to progressively replace the ambiguous depth estimations at the higher scale by reliable depths at the lower scale. As a result, we optimize both depth and probability maps in~\fref{Multi-metric}.

Considering a pyramid depth map $D^{k={0},\cdots,K-1}$ and a corresponding probability map $P^{k=0,\cdots,K-1}$ from VA-MVSNet,
we use the photometric consistency to measure the matching quality through the probability map and geometric consistency to measure the depth consistency between multiple images.
To select accurate and well-matched depth value in the lower scale $k+1$ depth maps, we only select the estimated depth which satisfies both the photometric and geometric consistency. 
Firstly, for the photometric consistency, we expect to iteratively replace unreliable depth values with low confidence $P^{k}(p)<\epsilon_{low}$ at the scale $k$ by reliable depths $P^{k+1}(p)>\epsilon_{high}$ at the downsampling scale $k+1$, where $P^{k}(p)$ denotes the confidence of pixel $p$ in the probability map $P^{k}$ and $\epsilon$ represent the filtering confidence threshold. 
After discarding mis-matched errors through the photometric consistency, we project a reference pixel $p$ of image $\textbf{I}_{i}$ to the corresponding pixel $p_{proj}$ in the neighbor image $\textbf{I}_{j}$ through $D_{i}(p)$ and camera parameters. In turn, we reproject $p_{proj}$ through $D_{j}(p_{proj})$ back to the reference image as $p_{reproj}$ with $d_{reproj}$. We remain the pixeles which satisfy the following geometric constraints in at least three neighbor views:
\begin{equation}
{\left\|p-p_{reproj}\right\|}_2<\tau_1, 
\end{equation}
\begin{equation}
{\left\|D_i(p)-d_{reproj}\right\|}_1<\tau_2\cdot D_i(p). 
\end{equation}

Through our multi-metric pyramid depth map aggregation, the reliable depths at a lower scale $k+1$ can be progressively propagated to replace the mismatched depths at $k$ scale until it leads to a final refinement at $k=0$ scale, which improves the robustness and completeness of 3D point cloud.



\section{Experiments}
\subsection{Implementation Details}\label{implementation_details}
\paragraph{Training} We train VA-MVSNet on the DTU dataset~\cite{aanaes2016DTU}, which consists of 124 different indoor scenes scanned by fixed camera trajectories in 7 different lighting conditions.
Following the common practices~\cite{huang2018deepmvs,ji2017surfacenet,yao2018mvsnet,yao2019recurrent,chen2019point}, 
we train our network on the training split and evaluate on the evaluation part and use the same depth maps provided by MVSNet~\cite{yao2018mvsnet}. 
During training, the input image size is set to $W\times H = 640\times 512$ and the number of input images $N=5$. The depth hypotheses are sampled from $425mm$ to $935mm$ with depth plane number $D=192$ in an inverse manner as illustrated in R-MVSNet~\cite{yao2019recurrent}. 
We implement our network on \textbf{PyTorch}~\cite{paszke2017automatic} and train it end-to-end for $16$ epochs using \textsl{Adam}~\cite{kingma2014adam} with an initial learning rate $0.001$ which is decayed by 0.9 every epoch. 
Batch size is set to $4$ on $4$ NVIDIA TITANX graphics cards.
\begin{table}[t]
    \centering
    \small
    \begin{tabular}{lccc}
    \hline
    \multirow{2}{*}{Method} & \multicolumn{3}{c}{Mean Distance (mm)} \\ 
    & Acc. & Comp. & \textsl{overall} \\ 
    \hline
    Colmap \cite{schonberger2016pixelwise} & 0.400 & 0.664 & 0.532 \\ 
    Gipuma \cite{galliani2015massively} & \textbf{0.283} & 0.873 & 0.578 \\
    MVSNet \cite{yao2018mvsnet} & 0.396 & 0.527 & 0.462 \\ 
    R-MVSNet \cite{yao2019recurrent} & 0.385 & 0.459 & 0.422 \\ 
    P-MVSNet \cite{luo2019p} & 0.406 & 0.434 & 0.420 \\
    PointMVSNet \cite{chen2019point} & 0.361 & 0.421 & 0.391 \\ 
    PointMVSNet-HiRes \cite{chen2019point} & 0.342 & 0.411 & 0.376 \\ 
     \hline
    \textbf{VA-MVSNet}  & 0.378  &  0.359 & 0.369 \\
    \textbf{PVA-MVSNet} & 0.379 & \textbf{0.336} & \textbf{0.357} \\
    \end{tabular}
    \caption{Quantitative results on the DTU evaluation dataset~\cite{aanaes2016DTU} (lower is better). Our VA-MVSNet and PVA-MVSNet (with voxel-wise view aggregation) outperform all methods in terms of completeness and overall quality with a significant improvement.}
    \label{eval_dtu}
\end{table}

\paragraph{Evaluation}
For testing, we use $N=7$ image views and $D=192$ for depth plane sweeping in an inverse depth setting. 
We evaluate our methods on \textsl{DTU} with an original input image resolution: $1600\times 1184$.
For \textsl{Tanks and Temples}, the camera parameters are computed by OpenMVG~\cite{moulon2014openmvg} following MVSNet~\cite{yao2018mvsnet} and the input image resolution is set to $1920\times1056$. We use the same multi-metric constraint parameters, where $\epsilon_{low}=0.5$, $\epsilon_{high}=0.9$, $\tau_{1}=1$ and $\tau_{2}=0.01$.

\begin{figure}[t]
  \centering
 \begin{tabular}{ccccccc}
    \includegraphics[width=0.158\columnwidth]{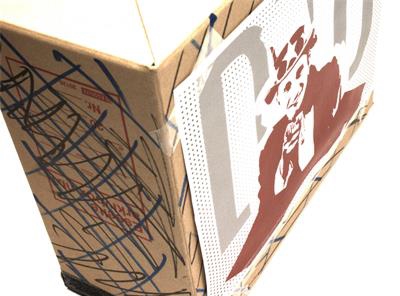}&
    \includegraphics[width=0.158\columnwidth]{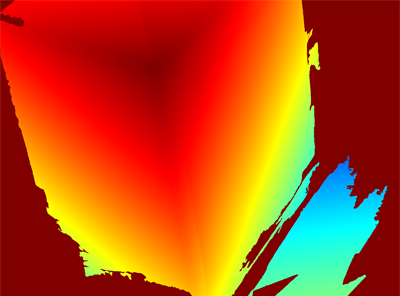}&
    \includegraphics[width=0.158\columnwidth]{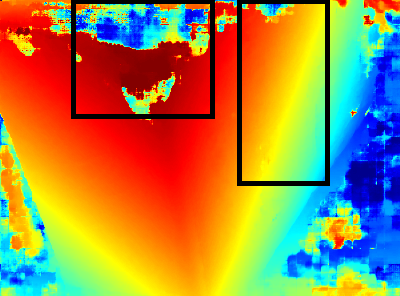}&
    \includegraphics[width=0.158\columnwidth]{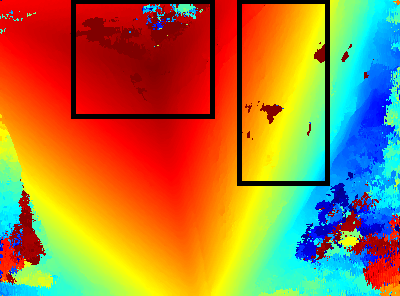}&
    \includegraphics[width=0.158\columnwidth]{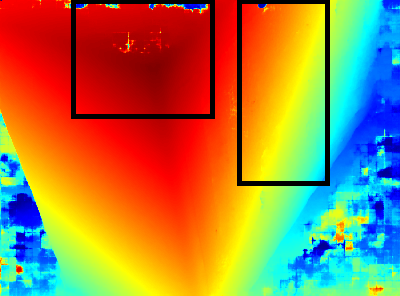}&
    \hspace{-0.005\columnwidth}\includegraphics[width=0.158\columnwidth]{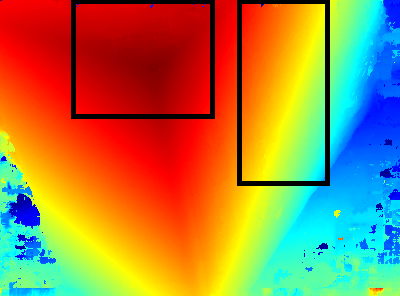}\\
    
      \includegraphics[width=0.158\columnwidth]{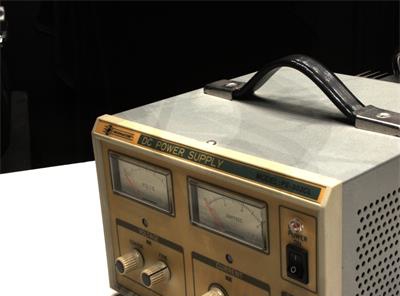}&
    \includegraphics[width=0.158\columnwidth]{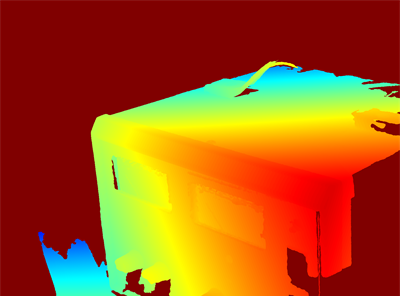}&
    \includegraphics[width=0.158\columnwidth]{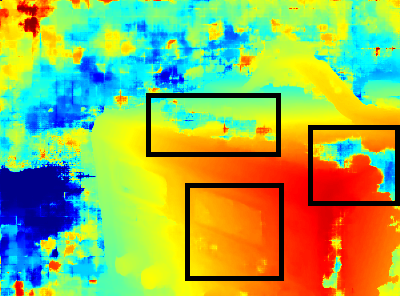}&
     \includegraphics[width=0.158\columnwidth]{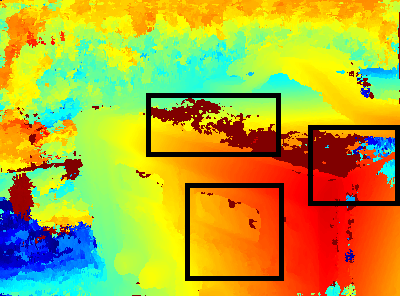}&
   \includegraphics[width=0.158\columnwidth]{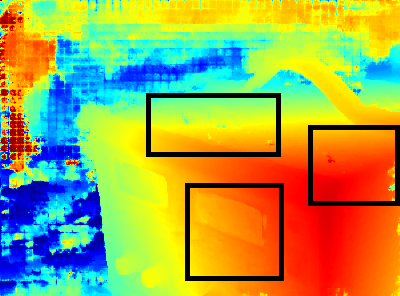}&
   \hspace{-0.005\columnwidth}\includegraphics[width=0.158\columnwidth]{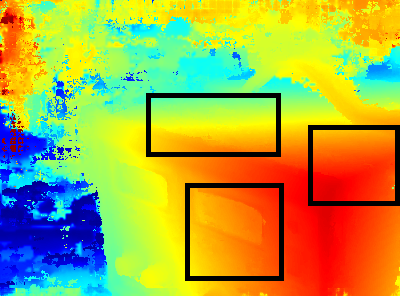}\\
   
   
    \hspace{-0.01\columnwidth}\scriptsize{a)Ref. image} & \hspace{-0.01\columnwidth}\scriptsize b)GT depth & \scriptsize c)MVSNet &  \hspace{-0.01\columnwidth} \scriptsize d)R-MVSNet & \hspace{-0.01\columnwidth} \scriptsize e)VA-MVSNet &\hspace{-0.01\columnwidth} \scriptsize f)PVA-MVSNet \\
 \end{tabular}
 \caption{Comparison of depth map estimations of \textsl{Scan $13$ and $11$} in the \textsl{DTU}~\cite{aanaes2016DTU}. Our VA-MVSNet and PVA-MVSNet achieve more accurate, continuous and complete depth map in comparison to ~\cite{yao2018mvsnet,yao2019recurrent} methods~\cite{yao2018mvsnet,yao2019recurrent}.}
 \label{dtu_cmp}
\end{figure}

\begin{figure}
    \centering
    \includegraphics[width=1.0\columnwidth]{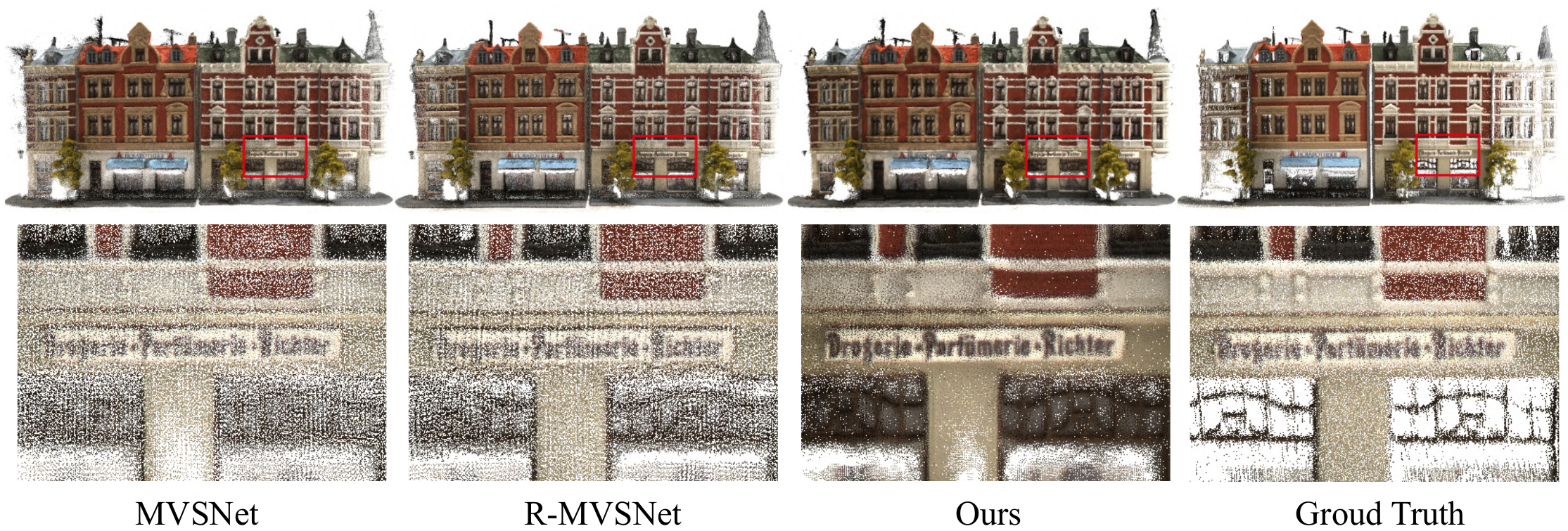}
    \caption{Comparison of reconstruction point clouds for the model \textsl{Scan 15} in the benchmark \textsl{DTU}~\cite{aanaes2016DTU}. Our method generates denser, smoother and more complete point cloud compared with other methods~\cite{yao2018mvsnet,yao2019recurrent}.}
    \label{dtu_point_cmp}
\end{figure}

\paragraph{Filtering and Fusion} 
We fuse all depth maps into a complete point cloud as in~\cite{galliani2015massively,yao2018mvsnet}. 
In our experiments, we only consider the reliable depth values with confidence larger than $\epsilon=0.9$ and utilize the aforementioned geometric consistency to select those pixels occurring in more than three neighbor views. Finally, the depths are projected to 3D space and fused to produce a 3D point cloud.

\subsection{Benchmarks Results}
\paragraph{DTU Dataset}
We evaluate our proposed method on the \textsl{DTU}~\cite{aanaes2016DTU} evaluation set. Quantitative results are shown in~\tref{eval_dtu}. The accuracy and completeness are calculated using the official matlab script provided by the \textsl{DTU}~\cite{aanaes2016DTU} dataset. 
The overall reconstruction quality is evaluated by calculating the average of the accuracy and completeness, as mentioned in~\cite{yao2018mvsnet,aanaes2016DTU}.
While Gipuma~\cite{galliani2015massively} performs the best regarding to accuracy, our PVA-MVSNet and VA-MVSNet establish a new state-of-the-art both in completeness and overall quality with a significant margin compared with all previous methods~\cite{tola2012efficient,yao2018mvsnet,yao2019recurrent,chen2019point}.
We compare our depth maps with~\cite{yao2018mvsnet,yao2019recurrent} in ~\fref{dtu_cmp}. VA-MVSNet predicts a more accurate, delicate and complete depth map by introducing different variances in multi-views through our proposed self-adaptive view aggregation. Moreover, PVA-MVSNet further fill-in the mismatched errors with reliable depths in the pyramid depth maps by our multi-metric pyramid depth map aggregation. Benefiting from more accurate, smooth and complete depth map estimation, our method can generate denser and more complete and delicate point clouds in~\fref{dtu_point_cmp}.

\begin{table*}[htbp]
    \centering
    \resizebox{\linewidth}{!}{
    \small
    \begin{tabular}{l|cccccccccc}
    \hline
    Method & Rank & Mean & Family & Francis & Horse & L.H. & M60 & Panther & P.G. & Train \\ \hline
    \scriptsize{MVSNet~\cite{yao2018mvsnet} }&	52.75&	43.48&	55.99&	28.55&	25.07&	50.79&	53.96&	50.86&	47.90&	34.69\\
    \scriptsize{R-MVSNet~\cite{yao2019recurrent}} &	42.62&	48.40&	69.96&	46.65&	32.59&	42.95	&51.88&	48.80&	52.00&	42.38\\ 
    \scriptsize{Point-MVSNet~\cite{chen2019point}} & 40.25&	48.27&	61.79&	41.15&	34.20	&50.79	&51.97&	50.85&	52.38	&43.06 \\ 
    \scriptsize{P-MVSNet~\cite{luo2019p}} &		\textbf{17.00} &		\textbf{55.62} &	\textbf{70.04} &		44.64 &		40.22 &		\textbf{65.20} &		55.08	 &	\textbf{55.17}	 &	\textbf{60.37} &	\textbf{54.29} \\ \hline
    \scriptsize{\textbf{PVA-MVSNet}} & 21.75 & 54.46 & 69.36 & \textbf{46.80} & \textbf{46.01} & 55.74 & \textbf{57.23} & 54.75 & 56.70 & 49.06\\
    \hline
    \end{tabular}
    }
    \caption{Quantitative Results on the \textsl{Tanks and Temples} benchmark~\cite{knapitsch2017tanks}. The evaluation metric is \textsl{f-score} which higher is better. (L.H. and P.G. are the abbreviations of \textsl{Lighthouse} and \textsl{Playground} dataset respectively. )}
    \label{tat_eval}
\end{table*}

\begin{figure}[t]
    \centering
    \includegraphics[width=0.98\textwidth]{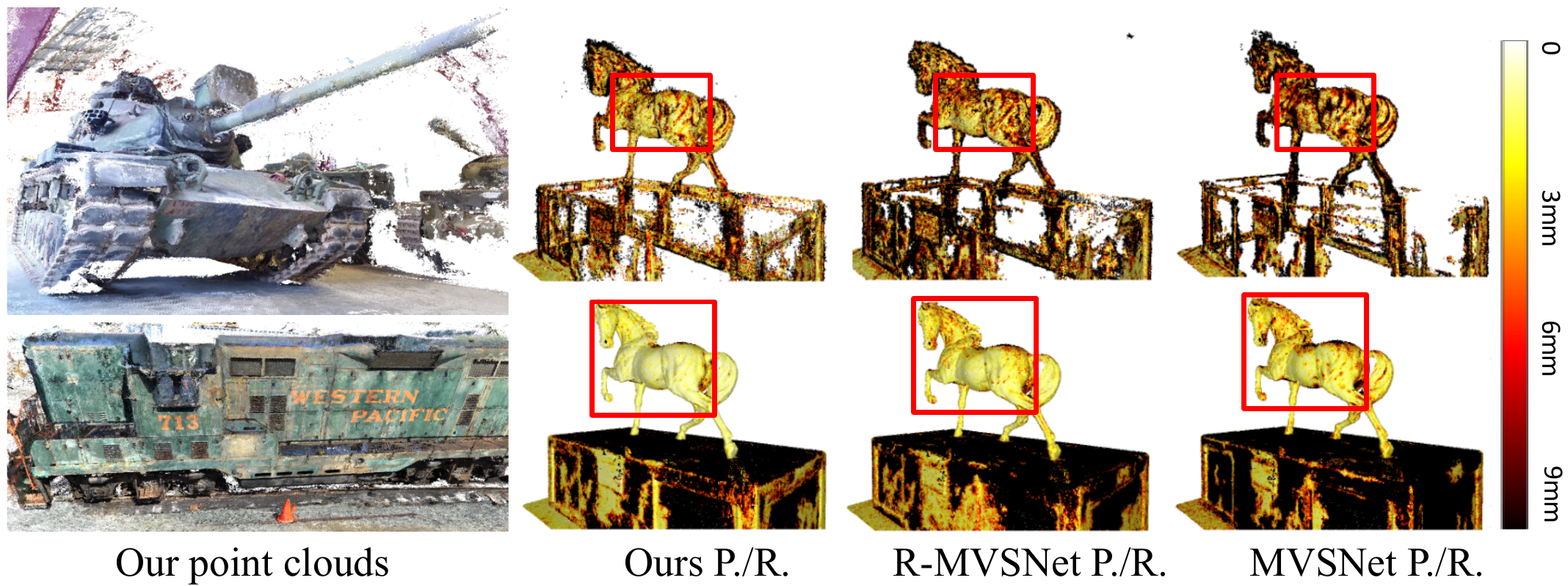}
    \caption{The visualization of our partial point cloud results and the comparison with~\cite{yao2018mvsnet,yao2019recurrent} on the Precision and Recall of \textsl{Horse} dataset on the \textsl{Tanks and Temples}~\cite{knapitsch2017tanks} benchmark. The darker means the bigger error.}
    \label{TNT_results}
\end{figure}

\paragraph{Tanks and Temples Benchmark}
To explore the generalization of PVA-MVSNet, we compare our method \textbf{without any fine-tuning} with other baselines~\cite{yao2018mvsnet,yao2019recurrent,chen2019point,ChenPMVSNet2019ICCV} on the \textsl{Tanks and Temples}, which is a more complicated outdoor dataset. \tref{tat_eval} summarizes the results.
The mean \textsl{f-score} increases from $43.48$ to $54.46$ (larger is better, date: Mar. 5, 2020) compared with MVSNet~\cite{yao2018mvsnet}, which demonstrates the efficacy and strong generalization of PVA-MVSNet.
Our method outperforms Point-MVSNet~\cite{chen2019point} significantly with a higher $13\%$ mean \textsl{f-score}, which is the best baseline on \textsl{DTU} dataset. And we achieve a comparable result with P-MVSNet~\cite{luo2019p}. 
The simple fusion process we adopted achieves a comparable result with P-MVSNet~\cite{luo2019p}, which uses an extra \textsl{refinenet} and more depth filtering process to pursue better performance.
Our partial reconstructed point clouds are shown in~\fref{TNT_results}, and we compare the Precision and Recall of the \textsl{Horse} dataset with~\cite{yao2018mvsnet,yao2019recurrent}, which is provided by the \textsl{Tanks and Temples}~\cite{knapitsch2017tanks} benchmark. Our method generates more accurate and complete point clouds with higher precision and recall than the others~\cite{yao2018mvsnet,yao2019recurrent}, due to the enhanced accuracy from self-adaptive view aggregation and the increased completeness and robustness from our multi-metric pyramid depth map aggregation.

\subsection{Ablation Studies}
In this section, we provide ablation experiments to analyze the strengths of the key components of our architecture. For following studies, to eliminate the non-learning influence, all experiments use the same consistency-check parameters in \sref{implementation_details} and are tested on the \textsl{evaluation} and \textsl{validation} \textsl{DTU}~\cite{aanaes2016DTU} dataset.

\makeatletter
\newcommand\figcaption{\def\@captype{figure}\caption}
\newcommand\tabcaption{\def\@captype{table}\caption}
\makeatother
\begin{figure}[t]
    \begin{minipage}[t]{.48\columnwidth}
    \centering
    \figcaption{Validation results of the mean average depth error with different components in VA-MVSNet during training.}
    \includegraphics[width=0.6\columnwidth]{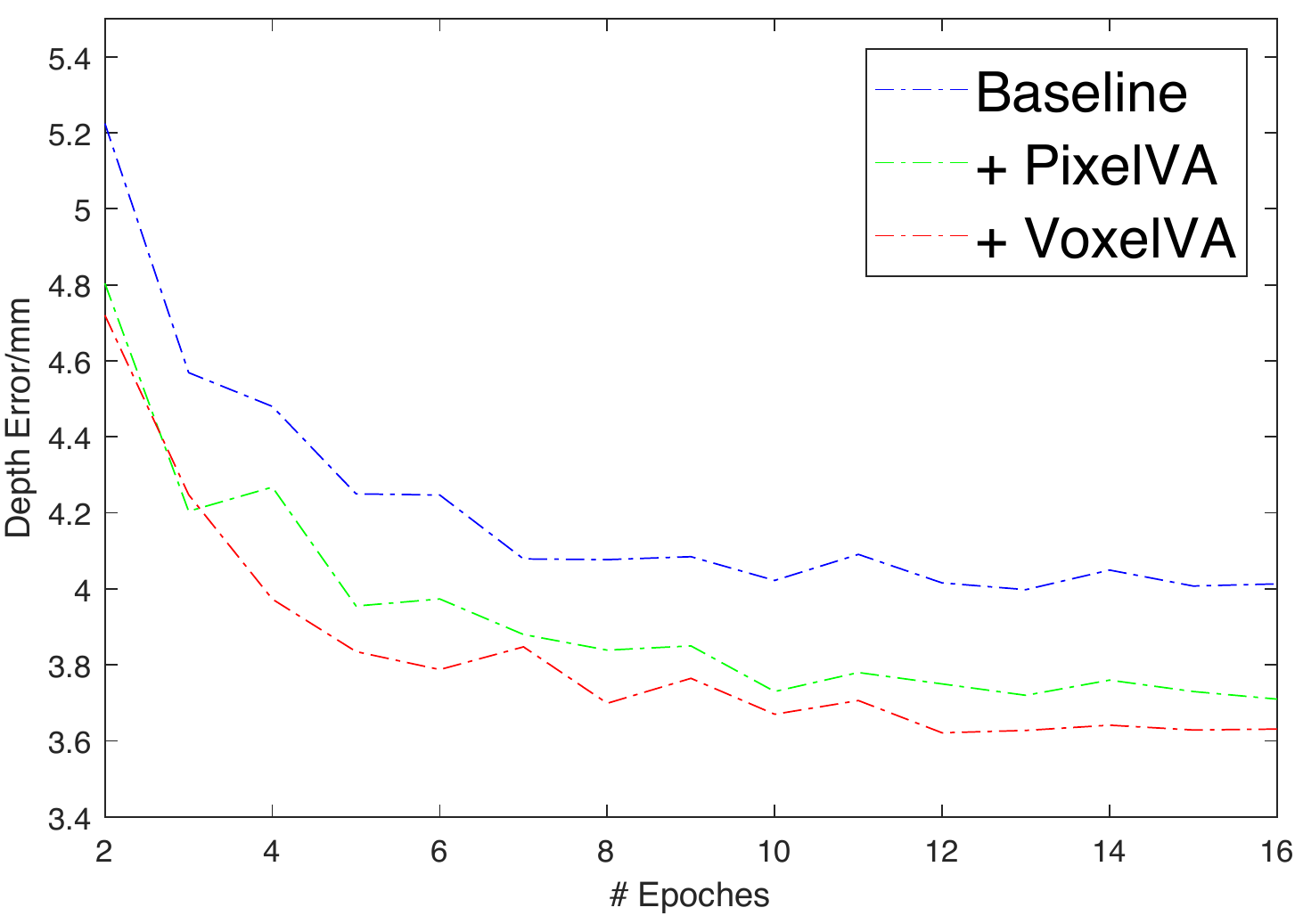}
    \label{depth_error}
    \end{minipage}
    \hfill
    \begin{minipage}[t]{.48\columnwidth}
    \centering
    \small
    \tabcaption{Contributions of different components in our architecture on the evaluation DTU~\cite{aanaes2016DTU}.} 
    \begin{tabular}{l|ccc}
    \hline
    Components & Acc. & Comp. & \textsl{Overall} \\ 
    \hline
    baseline & 0.454 & 0.372 & 0.413 \\
    +PixelVA & 0.390 & 0.369 & 0.379 \\
    +VoxelVA &  0.378 & 0.359 & 0.369 \\
    +PixelVA+MMP & 0.392 & 0.341 & 0.366 \\
    +VoxelVA+MMP & \textbf{0.379} & \textbf{0.336} & \textbf{0.357}   \\
    \hline
    \end{tabular}
    \label{abl_dtu}
    \end{minipage}
\end{figure}

\paragraph{Self-adaptive View Aggregation}
As shown in \tref{abl_dtu}, compared with our baseline method which is using the same mean square error as cost volume aggregation in MVSNet~\cite{yao2018mvsnet},
both PixelVA and VoxelVA can improve the results of 3D reconstruction point cloud with a significant margin, especially on the accuracy of reconstruction quality.
Specifically, the VoxelVA provides a $16.7\%$ increase on \textsl{accuracy}, which is better than the PixelVA $14.1\%$ due to the learning variance of the depth wise hypothesis. Besides, the VoxelVA has more parameters but less operations compared with PixelVA as denoted in \tref{weightnet}.
During training, as shown in~\fref{depth_error}, the depth error on \textsl{validation} dataset drops significantly by introducing our proposed novel self-adaptive view aggregation.




\begin{table}[t]
    \centering
    \resizebox{0.50\linewidth}{!}{
    \small
    \begin{tabular}{cc|c|ccc}
    \hline
    \multicolumn{2}{c|}{Number of views} & \multirow{2}{*}{\tabincell{c}{Number of\\ pyramid}} & \multirow{2}{*}{\tabincell{c}{Acc.\\ (mm)}} & \multirow{2}{*}{\tabincell{c}{Comp.\\ (mm)}} & \multirow{2}{*}{ \tabincell{c}{\textsl{Overall}\\ (mm)}} \\ 
    training & test & & & & \\
    \hline
    $N=3$ & $N=2$ & $\setminus$ &0.415 & 0.467 & 0.441 \\
    $N=3$ & $N=3$ & $\setminus$ &0.380 & 0.379 & 0.380 \\
    $N=3$ & $N=5$ & $\setminus$ & 0.381 & 0.361 & 0.371 \\
    $N=3$ & $N=7$ & $\setminus$ &0.380 &0.361 & 0.370 \\ \hline
    $N=4$ & $N=7$ & $\setminus$ & 0.380 & 0.359 & 0.370 \\ 
    $N=5$ & $N=7$ & $\setminus$ &0.378 & 0.359 & 0.369 \\ \hline \hline
    $N=5$ & $N=7$ & $K=1$ &\textbf{0.372} & 0.350 & 0.361 \\
    $N=5$ & $N=7$ & $K=2$ &0.378 & 0.341 & 0.360 \\
    $N=5$ & $N=7$ & $K=3$ &0.379 & \textbf{0.336} & \textbf{0.357} \\ \hline
    \end{tabular}
    }
    \caption{Ablation study on different number of views $N$ in training and testing phase and different numbers of image pyramid on \textsl{DTU}~\cite{aanaes2016DTU} evaluation dataset.} 
    \label{num_views_pyramid}
\end{table}

\paragraph{Number of Views}
We investigate the influence of variant numbers of views $N$ in different phases on \textsl{DTU} \textsl{evaluation} dataset. VA-MVSNet can process an arbitrary number of views and well leverage the variant importance in multi-views due to our proposed self-adaptive view aggregation. 
In the test phase, we use the model trained on $3$ views to compare the reconstruction results with different numbers of views $N=2,3,5,7$. As shown in \tref{num_views_pyramid}, the result with $N=5$ achieves a great improvement compared with $N=2, 3$, but the influence from two more extra views in $N=7$ is quite small which can be ignored. It demonstrates that our proposed self-adaptive view aggregation can well enhance the valid information in the good neighbor views and eliminate bad information in farrer views (the neighbor views are ranked by the matching quality with the reference view in SfM \cite{schoenberger2016sfm}). 
In the training phase, we compare the results on the input view $N=7$ using the models trained on $N=3,4,5$. The model trained on $N=5$ is slightly better than $N=3$ but with more training time.

\paragraph{Multi-metric Pyramid Depth Aggregation}
As shown in \tref{abl_dtu}, 
The completeness can be averagely improved by $7.0\%$ while introduce a negligible drop $0.39\%$, benefiting from ``MMP" (multi-metric pyramid depth aggregation).
As denoted in \textsl{e)} and \textsl{f)} in~\fref{dtu_cmp}, PVA-MVSNet improves VA-MVSNet by generating more delicate and complete depth maps. 
To better analyse the improvement from different pyramid level image, we explore the influence by different numbers of image pyramid in \tref{num_views_pyramid}. The $K=1$ level pyramid image improves both accuracy and completeness with a big margin. 
A trade-off between accuracy and completeness is achieved by using more pyramid images $k=2$ and $k=3$, it leads to reconstructed 3D point cloud with better overall quality.  



\begin{table}[t]
    \centering
    \resizebox{0.55\linewidth}{!}{
    \small
    \begin{tabular}{l|c|ccc}
    \hline
    Methods & H,W,D & Mem. & Time. & Overall \\ 
    \hline
    MVSNet & $1600,1184,256$ &  15.4GB & 1.18s & 0.462 \\ 
    R-MVSNet & $1600,1184,512$ &  \textbf{6.7GB} & 2.35s & 0.422 \\ 
    PointMVSNet & $1280,960,96$ & 7.2GB & 1.69s & 0.391 \\ 
    PointMVSNet-HiRes & $1600,1152,96$ & 8.7GB & 5.44s & 0.376 \\ \hline 
    VA-MVSNet  & $1600,1184,192$ & 18.1GB & \textbf{0.91s} & 0.369 \\
    PVA-MVSNet & $1600,1184,192$ & 24.87GB & 1.01s & \textbf{0.357} \\
    \hline
    \end{tabular}
    }
    \caption{Comparisons on the time and memory cost on the \textsl{evaluation} DTU~\cite{aanaes2016DTU} dataset. MVSNet and R-MVSNet are implemented in TensorFlow while others in PyTorch.}
    \label{time}
\end{table}

\subsection{Runtime and Memory Performance}
Given time and memory performance in \tref{time}, all methods are tested on GeForce RTX 2080 Ti. VA-MVSNet runs fast at a speed of $0.91s$ / view, even if it runs with the biggest memory consumption.
Unlike PointMVSNet~\cite{chen2019point}, multi-scale pyramid images can be processed independently in parallel. Therefore, with little extra time about $0.1s$ for multi-metric pyramid depth aggregation, the performance of 3D point cloud reconstruction increases significantly from $0.369$ to $0.357$ in PVA-MVSNet on \textsl{DTU}~\cite{aanaes2016DTU} dataset.

\section{Conclusion}
We present a novel pyramid multi-view stereo network with the self-adaptive view aggregation. The proposed VA-MVSNet dynamically selects the element-wise feature importance while suppresses the mismatching cost, which is quite efficient and effective. Casting in multi-scale pyramid images, benefiting from utilizing multi-metric constraint, PVA-MVSNet estimates a refined depth map for further improving the robustness and completeness of 3D reconstruction. 
Experimental results demonstrate that our proposed method PVA-MVSNet establishes a new state-of-the-art on the \textsl{DTU} dataset and shows great generalization by achieving a comparable performance as other state-of-the-art methods on \textsl{Tanks and Temples} benchmark without any fine-tuning.

\noindent\textbf{Acknowledgements} 
This project was supported by the National Key R\&D Program of China (No.2017YFB1002705, No.2017YFB1002601) and NSFC of China (No.61632003, No.61661146002, No.61872398).

%
%
\bibliographystyle{splncs04}
\bibliography{egbib}
\end{document}